%% file: main.tex
\documentclass[10pt,conference,twocolumn]{unoesc} 
\input{configuracoes/includes.tex}
\input{configuracoes/config.tex}
\usepackage{tikz}

\begin{document}

  \title{Sign Language to Text Conversion in Real Time using Transfer Learning}
  \author
  {
    \IEEEauthorblockN{Shubham Thakar, Samveg Shah, Bhavya Shah and Anant V. Nimkar\\}
    
    % \IEEEauthorblockN{Orientador(a): Nome Sobrenome\\}
    \IEEEauthorblockA{\textit{Department of Computer Engineering} \\
\textit{Sardar Patel Institute of Technology}\\
        Mumbai, India \\
    \{shubham.thakar, samveg.shah, bhavya.shah,anant\_nimkar\}@spit.ac.in}
  }

\maketitle

\input{secoes/01-resumo.tex}
\input{secoes/02-introducao.tex}

\input{secoes/03-instrucoes-gerais.tex}
% \begin{figure*}[h]
% \caption{CNN Architecture}
% \centering
% \includegraphics[width=0.8\paperwidth]{figuras/cnn architechture.png}
% \end{figure*}

% \clearpage
% \begin{tikzpicture}[h][remember picture, overlay, inner sep=0pt]
% \caption{CNN Architecture}
% \centering
% \node[anchor=north west] at (current page.north west) {\includegraphics[width=\paperwidth]{figuras/cnn architechture.png}};
% \end{tikzpicture}

% \begin{figure}
% \caption{CNN Architecture}
% \centering
% \includegraphics[width=1\paperwidth]{figuras/cnn architechture.png}
% \end{figure}

\input{secoes/04-CNN_architechture}

\begin{figure*}[h]

\centering
\includegraphics[width=0.7\paperwidth]{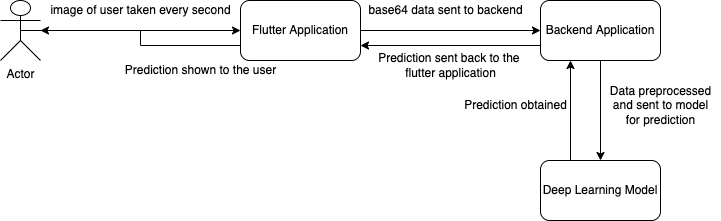}
\caption{Sign language to text conversion application design}
\end{figure*}

\input{secoes/05-Methodology.tex}

\input{secoes/06-Results.tex}
\input{secoes/07-conclusao.tex}
\input{secoes/08-futurescore}

\input{secoes/09-ref.tex}
\end{document}

%% file: configuracoes/includes.tex
\usepackage[%usado para determinar medidas
top=1.78cm,
bottom=1.78cm,
left=1.65cm,
right=1.65cm,
headsep=0cm,
]{geometry}
\usepackage{times}
\usepackage{enumitem}%redefinir espacos itemize
\usepackage{graphicx}
\usepackage{url,hyperref}
\usepackage[utf8]{inputenc}
\usepackage{float}%mais controle para manipular figuras
\usepackage{caption}%manipular legenda da figura e tabela
\usepackage{mathtools}%equacoes
\usepackage[hang,flushmargin]{footmisc} 
\usepackage{xcolor}
\usepackage{wrapfig} %usado para envolver figura com texto
\usepackage{fancyhdr}%criacao do cabecalho
\usepackage{etoolbox}
\usepackage[export]{adjustbox}%mais controle para ajustar tamanho da tabela
\usepackage{comment}%ambiente para comentario
\usepackage{relsize} %usado por comandos \mathlarger
\usepackage{lipsum} 

\usepackage[
    style=numeric,
    sorting=none,
    maxbibnames=10]{biblatex}
\usepackage[english]{babel}

%% file: configuracoes/config.tex
\makeatletter
\newcommand{\linebreakand}{%
  \baselineskip 
  \end{@IEEEauthorhalign}
  \hfill\mbox{}\par
  \mbox{}\hfill\begin{@IEEEauthorhalign}
}
\makeatother

\addto\captionsbrazil{
  \renewcommand{\abstractname}{Abstract}
  \renewcommand{\tablename}{Tabel}
}

\addto\captionsenglish{
  \renewcommand{\tablename}{TABLE}
}

%% file: secoes/01-resumo.tex
 \vspace{0.3cm}
  \begin{abstract}
  \noindent The people in the world who are hearing impaired face many obstacles in communication and require an interpreter to comprehend what a person is saying. There has been constant scientific research and the existing models lack the ability to make accurate predictions. So we propose a deep learning model trained on the ASL i.e the American Sign Language which will take actions in the form of ASL as input and translate it into text. To achieve the translation a Convolution Neural Network model and a transfer learning model based on the VGG16 architecture are used. There has been an improvement in accuracy from 94\% of CNN to 98.7\% by Transfer Learning an improvement of 5\%. An application with the deep learning model integrated has also been built.
  \end{abstract}

  
    \vspace{0.5cm}

%% file: secoes/02-introducao.tex
\section{Introduction}

As per the World Health Organization, there are over 1.5 billion people who live with hearing loss and require rehabilitation to address their 'disabling' hearing loss. The American Sign Language(ASL) is one of the most popular languages used by people who are deaf to communicate with each other. With its natural syntax, which shares the same etymological roots as spoken languages but has a different grammar, ASL may express the outcome of bodily acts. It becomes difficult for people not familiar with ASL to communicate with people who are hearing impaired. 
 \vspace{0.3cm}

ASL[17] is a full, natural language with English-like syntax that shares many of the same linguistic characteristics as spoken languages. Hand and face gestures are used to convey meaning in ASL. The current technologies existing for sign-to-speech conversion include wearable which requires physical hardware devices to achieve the same. There are also technologies existing that depend on hand detection using a leap motion sensor. These technologies have a high hardware requirement and accuracy of about 70\%. There are many models which have been built using LSTM-RNN, lightweight CNN, etc. 
\vspace{0.3cm}

We propose using a deep learning model as a translator that can convert sign language to text. Three convolutional layers in all were used in the CNN model to extract features. The number of kernels was set to 32 for the first convolutional layer and to 64 for the following two layers. The model comprises two thick layers. The output space's dimensionality in the first layer using the ReLu activation function is set to 128. Since the VGG16[18] model utilized represented the outcome from the ImageNet dataset, it did not have the 1*1*1000 topmost layer. After flattening the second-to-last layer of the VGG16 model, two further layers were added. The photos were reduced in size to 64 by 64 in order to expedite training while preserving their great nuances. The data were scaled down between the values 0 and 1 after they had been imported into the NumPy array in order to prevent the exploding gradients problem that is often present in CNN and transfer learning. 

\vspace{0.3cm}

The entire model has been converted into an application that involves the user's action being captured by his/her camera and the data will be sent to the rest API in base64 format. Each YUV channel would include a user image that would be taken once every second. These YUV channels were changed from YUV to RGB and then into base64 strings. These base64 strings were transmitted via the rest API to a Django backend. The base64 string will first be transformed to an RGB picture on this backend. It will then be reduced in size to a 64*64 picture and transformed into a NumPy array. The numbers in this NumPy array will be scaled down to the range of 0 to 1 by multiplying them by 255. The transfer learning model will receive this data and produce the necessary prediction. The program will then get this forecast and display it to the user.
 \vspace{0.3cm}

Section II describes the literature survey undertaken about various deep learning models used to achieve sign-to-text conversion. In section III we have described our first approach which is the CNN architecture. Further in section IV, we have described our methodology in detail which includes transfer learning. Section V talks about our application which we built using the deep learning model followed by section VI where we describe the results. In section VII and section VIII, we talk about the conclusion and the future scope.

%% file: secoes/03-instrucoes-gerais.tex
\section{Literature Survey}
This section discusses the existing research work in the domain of sign language recognition and translation.

\vspace{0.3cm}
%CNN and transfer learning
Convolutional Neural Network (CNN) is the most common technique used for ASL recognition. Hsien-I Lin et al. [1] have used image segmentation to extract the hand from the image. This was then followed by modeling the skin and then to center the image around the primary axis, they calibrate the threshold image around it. To train and forecast the results of a convolutional neural network model, they used this image. Their algorithm was trained on seven hand motions, and when applied to those movements, it delivers an accuracy of about 95\%. The authors [2] have used the American Sign language dataset and built a real-time system to translate it into text. They capture the images from a webcam and perform a hand gesture scan as preprocessing of data to make sure the gesture has been amplified. Then it is followed by feeding the data to the Keras CNN model. The model that has already been trained produces the projected label. Every label for a gesture has a probability attached to it. The label with the highest possibility becomes the anticipated label. Then the classifiers are used to predict the final label. The accuracy achieved overall for the same is 95.8\%. Garcia, B et al. [3] have attempted to create a real-time sign language translator which can help facilitate communication between the deaf community and the general public. They used a pre-trained GoogLeNet architecture for the classification task. Their model was able to classify letters a-e correctly with first-time users. Another real-time sign language translation technique was proposed by S. S Kumar [4]. This approach utilized time series neural networks for sign language conversion.
\par

%Light weight models
Some authors have tried to develop lightweight models for ASL classification.
 [5], [6] attempt to create a lightweight CNN architecture. Their aim was to scale down the architecture of the deep network to reduce the computational cost. For this Efficient Network (EfficientNet) models and generated lightweight deep learning models were utilized. Authors [7] have tried to create an app for the deaf community. But creating an app is not an easy chore because it calls for numerous efforts, such as memory usage and a flawlessly refined design. Their software takes a photograph of a sign motion and afterward transforms it into a relevant word. To begin with, they compared the gesture using a histogram that was connected to the sample test and more samples that are required to BRIEF in order to essentially lighten the load on the CPU and its processing time. They have outlined the steps necessary to add a gesture to their software and store it in their database so that they can increase their detection set in the future.

% New techniques
Various other techniques such as the use of Hidden Markov Models and LSTM-RNN have also been tried.
V. N. T. Truong et al. [8] have integrated the principles of AdaBoost and Haar-like classifiers. The dataset under consideration here is ASL which is the American Sign Language. A feature they have used is a big dataset to improve the accuracy of the model. They used a dataset of 28000 positive images coupled with 11100 samples of Negative images were used to implement and train the translator. A camera was used to capture data and send it to the model for processing. Another concept that has been used is HMM which is the  Hidden Markov model [9]. The dynamic features of gestures are addressed in this approach. There has been tracking of skin color blobs corresponding to the hands into a body–facial space centered on the user's face to extract gestures from a succession of video images. Differentiating between deictic and symbolic movements serves as the goal. The image is filtered using an indexing table with quick lookups. After filtering, pixels with similar skin tones are clustered into blobs. Based on the position (x, y) and colorimetry (Y, U, V) of skin color pixels, blobs are statistical objects that help identify homogeneous areas. LSTM-RNN [10] [11] is another approach that has been used for ASL classification. The former has proposed the kNN method for recognizing 26 alphabets. Angles between fingers, sphere radius, and distance between fingers positions are some features that were extracted for the classification model.

%Other languages
Various sign languages such as Arabic, and Indian also exist. Researchers have tried to build models to recognize them and correctly classify them. Saleh, Yaser, et al. [12] have tried to recognize Arabic sign language by fine-tuning deep neural networks. The authors have used a modification of the ResNet152 model structure for classifying hand gestures. A technique to translate Malayalam into Indian sign language known as a synthetic animation producing approach is discussed in [13]. HamNoSys is utilizing the intermediate representation in this way for sign language. With this technique, the application accepts some word sets, say one or more, and creates an animated section from them. A system that is interactive further turns the words into the HamNoSys-designed structure. The Kerala government uses its program to parse everything that has been developed to teach sign language and subtle awareness. [14] is entirely based on the Spanish speaking language, which converts the basic words into Spanish. This is advantageous for Spanish-speaking deaf people because it will give them a position to understand the sign language at a faster rate because it will be converted into the Spanish language rather than into English, which is frequently used as ASL. The tool or program they created for this consists of a variety of terms, including convertor to speech, which essentially turns those whole bits into a meaningful sentence in Spanish, and of course, a visible interface, which is used by the deaf person to specify the sequence of sign data, and translator, which merely converts those series in Spanish based language in a formed series.
  
%Other techniques
D. Kelly et al. [15] have created a system that operates continuously and uses a sequence of sign language gestures to create an automated training set and deliver the spots signs from that set. They have put forth a system that supervises the sentence and determines the associated compound sign gesture using the supervision of noisy texts, using instance learning as a density matrix technique. The group that was first intended to demonstrate the continuous data stream of words is now used as a training group for identifying gesture posture. They have experimented with this small sample of automated data that is used for their training, identification, and storage of subtle sign data. There are about 30 sign language files preserved. A number of experiments [16] to develop a statistical model for deaf persons to convert speech data into sign language were conducted. With the aid of an animated presentation and a statistical translation module for several sets of signs, they have further developed a system that automates speech recognition via ASR. As they continued, they employed the phrase-specified system and the state transducer for the translation process. According to the evaluation, specific figure types have been used: WER, BLEU, and then NIST. This paper illustrates the method for voice translation using an automation recognizer with all three of the aforementioned setups. The output of the finite type state transducer used in this study has a word error rate that ranged between 28.21 percent and 29.27 percent.

%% file: secoes/04-CNN_architechture.tex
\section{CNN Architecture}
This section discusses the CNN approach that was used to solve the problem of ASL classification. Section 5 further compares the results obtained using the CNN approach and transfer learning approach. \par
A Convolutional Neural Network, also known as CNN or ConvNet, is a class of neural networks that is adept in processing data that has a grid-like topology, such as an image. 
The three layers of a CNN are typically convolutional, pooling, and fully connected layers. \par
The model had a total of 3 convolutional layers for feature extraction. For the 1st convolutional layer, the number of kernels was set to 32, for the other 2 layers the number of kernels was set to 64. Kernel size for all convolutional layers was 3*3 and the stride was set to 1. We used the ReLU activation function in all convolution layers. \par
Our model has 3 max-pooling layers one after each convolutional layer. The pool size for each layer was set to 2*2. \par
The model has 2 dense layers. In the first layer with the ReLu activation function, the dimensionality of the output space is set to 128. The second layer which has a softmax activation function has dimensionality equal to the number of classes for classification.

 \vspace{0.3cm}

%% file: secoes/05-Methodology.tex
\section{Deep Learning Model For Converting Sign Language To Text}
The huge advantage of deep learning models for sign language to text conversion is that different levels of visual information are processed on each layer. Lower layers process (detect) very local features, such as small sections of curves. The higher you go, the more complex the features become. And you can still interpret the network's functionality reasonably well. To adapt a model to a new purpose, the lower layers are usually kept and only the higher ones are trained to infer the features for the specific case. This significantly accelerates the training.
 \vspace{0.3cm}
 
\subsection{Processing Of The Images }
\begin{figure}[!htb]
\centering
\includegraphics[width=0.5\textwidth]{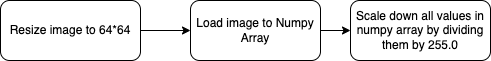}
\caption{Steps for pre-processing of the images}
\end{figure}
To speed up training while maintaining the fine details of the photos, the images were downsized to 64 by 64. To avoid the exploding gradients problem typically found in CNN and transfer learning, the data were scaled down between the values 0 and 1 once they had been loaded into the NumPy array. The steps for processing the image are depicted in Fig. 2.

\subsection{Dataset}
The ASL (American Sign Language) dataset which consisted of 87000 images was used to train and test the deep learning sign language to text conversion model. This dataset has 29 labels ( A …. Z, space, delete, nothing) and for the training purpose letter A corresponds to 0, B corresponds to 1 … , nothing corresponds to 28.

To speed up training while maintaining the fine details of the photos, the images were downsized to 64 by 64. To avoid the exploding gradients problem typically found in CNN and transfer learning, the data were scaled down between the values 0 and 1 once they had been loaded into the NumPy array. The steps for processing the image are depicted in Fig. 2.

\subsection{Model Architecture}
% \begin{figure}[h]

% \centering
% \includegraphics[width=0.5\textwidth]{figuras/vgg.png}
% \caption{VGG16 Architecture}
% \end{figure}
The model used for sign language to text conversion is based on transfer learning. A pre-trained model serves as the basis for a new model in transfer learning. Simply said, a model developed for one work is applied to another, related task as an optimization to facilitate quick modeling progress on the first task. Performance-wise, transfer learning models beat traditional deep learning models. This is because the models that include data (features, weights, etc.) from previously trained models already possess a comprehensive grasp of the features. Compared to creating neural networks from scratch, it speeds up the process. We decided to apply transfer learning to identify sign language using the VGG16 model that was previously trained on the ImageNet dataset. VGG16 was selected because it has an object identification and classification algorithm that can accurately categorize 1000 pictures into 1000 different groups with a 92.7 percent accuracy rate. It is a popular method for categorizing photographs and is easy to use with transfer learning. The VGG16 model used did not have the 1*1*1000 topmost layer, since it represented the result from the ImageNet dataset. The second last layer of the VGG16 Model was flattened and then 2 layers were added to the VGG16 model :
\begin{itemize}
    \item Dense layer with 512 Units. It has the sigmoid activation function. 
    \item Dense layer with 29 Units. It has the softmax activation function.
\end{itemize}
The outcome has the best probability of being the unit with the highest value in the final layer. The alphabet corresponding to this unit is the expected output. 
The model architecture is depicted in Fig. 3
\begin{figure}[h]

\centering
\includegraphics[width=0.5\textwidth]{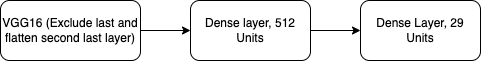}
\caption{Transfer learning model architecture}
\end{figure}

\subsection{Model Training}
The model is trained on a dataset of 29 different labels. The entire dataset was divided into two categories test data and train data. The train data had 80000 images and the test data had 7000 images. The total trainable parameters were 15,778,653. The training was done with a batch size of 128 and a learning rate of 0.0001 - 0.0002. Because it offers an optimization approach that can manage sparse gradients on noisy problems, the adam optimizer was used. The training lasted for 20 epochs. 

\section{Sign language to text conversion application}

Flutter was utilized to build the application. The screen orientation of the app was horizontal to facilitate the user to make the gestures and get the predictions on the same screen.   Fig. 4 represents a screenshot of the application converting sign language to text.  User pictures would be captured every 1 second in the form of YUV channels. These YUV channels were converted into RGB channels and transformed into base64 strings. These base64 strings were sent to a Django backend via rest API. On this backend, first, the base64 string will be converted to an RGB image. Then it will be resized to a 64*64 image and converted into a NumPy array. This NumPy array will be scaled down from 0 to 1 by dividing the numbers in the array by 255. This data will be fed to the transfer learning model which will generate the required prediction. This prediction will then be returned back to the application which will show it to the user. Fig. 1 represents the design of the application.

\begin{figure}[h]
\centering
\includegraphics[width=0.5\textwidth]{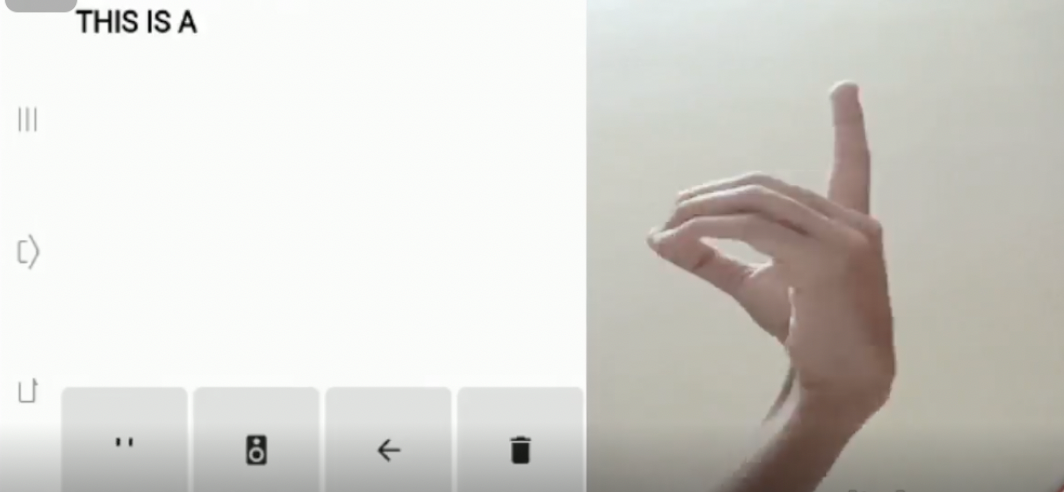}
\caption{Application screenshot}
\end{figure}

%% file: secoes/06-Results.tex
\section{Results and Discussions}
Cross-validation was used to test whether our model was overfitting, which is when the model fits the training data well but is unable to generalize and produce accurate predictions for new data. The data was split into two sets: the training set and the validation set. The model was trained using the training set, and its performance was assessed using the validation set.
On the validation set, the model has a 98.7 percent accuracy rate, while on the training set, it has a 98.8 percent accuracy rate. According to this, the model should operate with an accuracy of 98.7\% on new data. The model has a loss of 0.35\%. It can be observed that in Fig 5, the train and validation loss reduced with an increase in the number of epochs. Also, from Fig 6, it can be observed that the train and validation accuracy increased with the number of epochs.

\begin{figure}[h]

\centering
\includegraphics[width=0.5\textwidth]{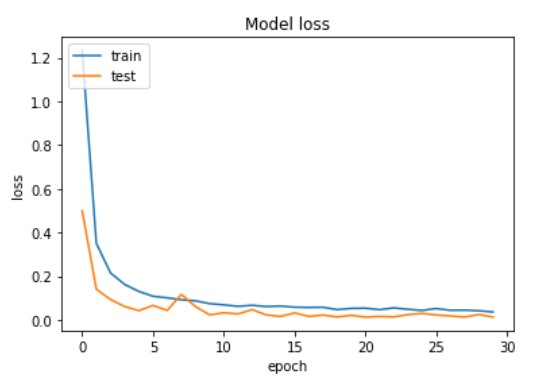}
\caption{Train and validation loss reduces with increase in epochs}
\end{figure}

\begin{figure}[h]

\centering
\includegraphics[width=0.5\textwidth]{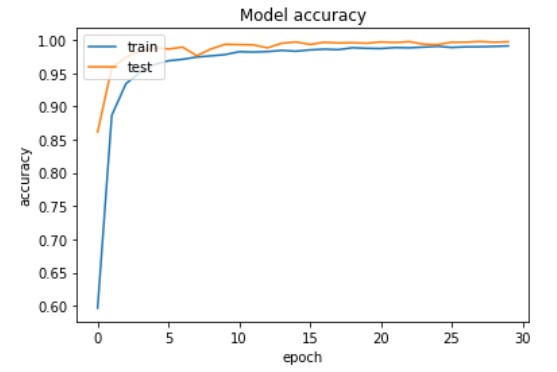}
\caption{Train and validation accuracy increases with increase in epochs}
\end{figure}

%% file: secoes/07-conclusao.tex
\section{Conclusion}
In this paper, we have presented a deep learning model which gave a high accuracy based on the concept of Transfer Learning and compared it with the existing CNN model and showed significant improvement. We pointed out existing loopholes in the deep learning models existing and built a new model. The model shown in this paper was able to achieve an accuracy of 98.7\% an improvement of over 4\%. This was possible because of the use of transfer learning with VGG16 along with Imagenet weights. In addition, obtaining the optimal hyper-parameters after rigorous experimentation also added to the accuracy of the model. The image needs to be resized to 64*64 pixels. We have further proposed an architecture of an application that is built for easy use of the model.
%Exemplo de comentario
\begin{comment}
Aqui também é um comentário
\end{comment}

%% file: secoes/08-futurescore.tex
\section{Future Scope}
This deep learning model is trained over the American sign language dataset, we can try to diversify this model by training over the Indian sign language dataset, British sign language dataset, etc. In addition to this, we assumed that the image on which the prediction will be made will have a smooth background, this can be improved by adding some noise to the image using augmentation techniques.